\documentclass{article}

\usepackage[preprint]{corl_2026} 

\usepackage{graphicx}
\graphicspath{{paper/figures/}{paper/}{figures/}{./}{../figures/}}
\usepackage{amsmath, amssymb}
\usepackage{booktabs}
\usepackage{tabularx}
\usepackage{wrapfig}
\usepackage{placeins}
\usepackage{caption}
\usepackage{subcaption}
\newcolumntype{Y}{>{\raggedright\arraybackslash}X}

\title{%
  \includegraphics[height=1.0cm,keepaspectratio]{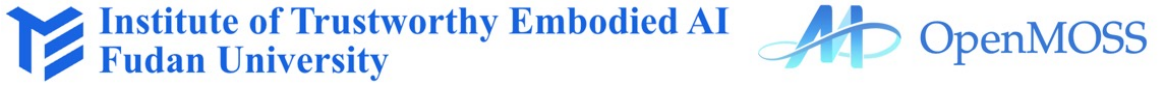}\\[0.8em]
  Coarse-to-Control: Action-Token Planning for Vision-Language-Action Models
  }
\author{%
\normalfont
Jinhao Wu$^{1,2}$ \quad
Shiduo Zhang$^{2,3}$ \quad
Yicheng Liu$^{4}$ \quad
Xiaopeng Yu$^{2,3}$ \quad
Sixian Li$^{2,3}$\\
Siyin Wang$^{2,3}$ \quad
Hang Zhao$^{4}$ \quad
Jing Huo$^{1}$ \quad
Yang Gao$^{1}$ \quad
Jingjing Gong$^{2,*}$ \\
Xipeng Qiu$^{2,3}$ \quad
Yu-Gang Jiang$^{3,*}$ \\
$^{1}$Nanjing University  \quad
$^{2}$Shanghai Innovation Institute \quad
$^{3}$Fudan University \\
$^{4}$Tsinghua University\quad
}

\begin{document}

\maketitle

\begingroup
\renewcommand{\thefootnote}{\fnsymbol{footnote}}
\footnotetext[1]{Corresponding authors.}
\endgroup

\begin{abstract}
Most vision-language-action (VLA) models map observations directly to actions without explicit intermediate planning,
which limits performance on long-horizon tasks where early mistakes compound. We propose \emph{Coarse-to-Control}, a
plan-execute VLA that introduces planning natively in the action-token space. The key idea is to let the policy first
predict a compact sequence of coarse action tokens that summarize the intended future trajectory, and then generate
executable action tokens conditioned on this plan. Because both planning and execution share a unified discrete action
vocabulary, the plan stays close to the control manifold and provides directly actionable guidance rather than an
abstract hint that must be translated back to motor commands. Experiments on LIBERO, SimplerEnv-WidowX, and real-world
manipulation tasks show that action-token planning consistently improves over direct action generation, with the
largest gains on long-horizon multi-stage tasks.
\end{abstract}

\keywords{Vision-Language-Action Models, Robot Learning}

\section{Introduction}
Vision-language-action (VLA) models have made rapid progress by learning to map visual observations and language
instructions directly to robot actions
\citep{brohan2022rt1,brohan2023rt2,ghosh2024octo,kim2024openvla,black2024pi0}.
Yet this direct-generation paradigm faces a fundamental representational tension: language instructions specify
\emph{what} to accomplish---``pick up the cup,'' ``place the carrot on the plate''---but say nothing about \emph{how}
to move. To generate precise motor commands, the policy must silently resolve approach direction, wrist orientation,
grasp pose, and waypoint sequence from a high-level goal description. Without an explicit intermediate layer that
bridges semantic intent and motor detail, all of this resolution is compressed into a single forward pass, asking the
policy to operate at two disparate levels of abstraction simultaneously.

Humans avoid this problem because the brain maintains distinct levels of motor representation. Research on skilled
movement suggests that action is organized hierarchically: the nervous system first specifies a high-level plan (goal,
movement direction, and grasp configuration), then refines execution through lower-level motor commands under continuous
sensory feedback \citep{rosenbaum2009human,bernstein1967co,lashley1951problem}. Crucially, this intermediate
representation is not a language description. The instruction \emph{place the carrot on the plate} conveys the semantic
goal but encodes nothing about arm trajectory, wrist orientation, or gripper timing. A coarse motor plan carries spatial
and temporal structure that language abstracts away; providing this structure explicitly, rather than leaving it
implicit in the goal description, is what enables more consistent and precise execution.

Recent reasoning-augmented VLAs address this gap by inserting intermediate representations before action generation:
textual rationales \citep{zawalski2024ecot,huang2025thinkact}, predicted visual subgoals
\citep{zhao2025cotvla,zhang2025dreamvla}, or spatial and structured reasoning representations
\citep{qu2025spatialvla,gao2025vlaos,huang2025graphcotvla}. Each improves task understanding, but none fully bridges the
underlying mismatch: text and images operate at a semantic or perceptual level rather than the level of motor intent,
leaving the policy to still infer the spatial and temporal structure that precise execution requires.

This motivates a planning medium that is more control-aligned than text or images yet more structured than direct action
prediction: one that lives in the action space itself. We propose \emph{Coarse-to-Control}, a plan-execute VLA that
first predicts coarse planning tokens summarizing the intended future trajectory and then generates executable action
tokens conditioned on this plan. The key enabler is a joint plan-execute tokenizer that maps both planning and execution
into a shared residual-VQ action vocabulary, so the plan remains close to the control manifold and provides more
actionable guidance than an abstract hint that must be translated back to motor commands.

The policy is trained end-to-end from demonstrations with the same autoregressive objective used for action prediction:
planning tokens are generated as an internal prefix, and executable tokens are generated conditioned on that prefix.
At inference, only executable tokens are decoded into robot actions, keeping planning as lightweight latent guidance
without a separate planner-controller interface. Coarse-to-Control reaches 97.90\% average success on LIBERO and 83.3\%
on SimplerEnv-WidowX, and improves robustness across four real-world manipulation tasks.

Our contributions are:
\begin{itemize}
  \item We identify that generating precise motor commands benefits from an intermediate motor-level intent, and argue
  that the action space is a more natural medium for this representation than text or images.
  \item We propose \emph{Coarse-to-Control}, a plan-execute VLA with a joint tokenizer that places coarse planning
  tokens and executable action tokens in a shared discrete vocabulary, so the plan stays close to the control manifold
  and provides more actionable guidance than text or image intermediates.
  \item We demonstrate on LIBERO, SimplerEnv-WidowX, and real-world manipulation tasks that action-token planning
  consistently improves over direct action generation and outperforms textual, visual, and spatial CoT baselines.
\end{itemize}

\section{Related Work}

\paragraph{Vision-Language-Action Models.}
VLA models adapt pretrained vision-language backbones to robot control by learning to output actions from multimodal
observations and language instructions \citep{brohan2022rt1,brohan2023rt2,ghosh2024octo,kim2024openvla}. Recent systems
such as $\pi_0$, $\pi_0$-FAST, and OpenVLA-OFT improve generality, action-token efficiency, and
speed-success trade-offs \citep{black2024pi0,pertsch2025fast,kim2025openvlaoft}. We build on these
action-generation backbones and add explicit planning before execution.

\paragraph{Intermediate Reasoning for VLAs.}
Recent VLAs introduce intermediate reasoning before control, including textual \citep{zawalski2024ecot,black2025pi05,huang2025thinkact}, visual-linguistic \citep{zhao2025cotvla, zhong2026dualcotvla}, 3D-aware spatial \citep{qu2025spatialvla,huang2025graphcotvla}, and trajectory-based reasoning \citep{duan2025molmoact,fang2026molmoact2}, as summarized in Figure~\ref{fig:reasoning_representation}.
Action CoT is closest to ours in moving reasoning toward motor outputs \citep{zhong2026acotvla}. We instead represent reasoning as a coarse future trajectory in the same executable action-token space used for control, directly unifying planning and execution.

\paragraph{Action Tokenization.}
Discrete action representations make continuous robot control compatible with sequence models. Existing methods primarily treat action tokens as compact representations for policy execution, using per-dimension binning \citep{brohan2023rt2,kim2024openvla}, DCT-based compression \citep{pertsch2025fast}, or neural tokenization \citep{liu2025faster,dong2026actioncodec,wang2025vq} to balance reconstruction fidelity, compression efficiency, and structural expressiveness. Our contribution is complementary: through a joint tokenizer shared across both stages, we use action tokens not only for low-level execution, but also as an interface for high-level planning.

\section{Coarse-Action Planning as Chain-of-Thought}

\begin{figure}[!htbp]
\centering
\includegraphics[width=\linewidth]{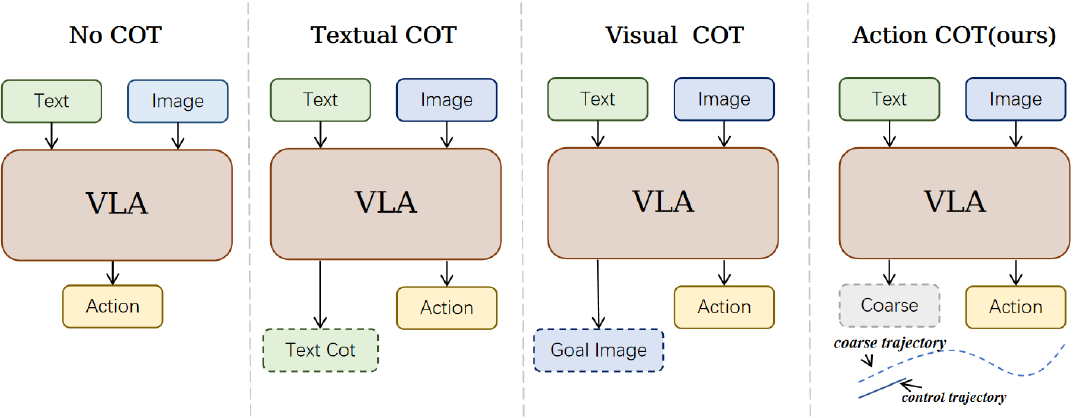}
\caption{Comparison of reasoning paradigms for VLA control: No CoT, textual CoT, visual CoT, and our action-token CoT. Our method uses a compact action-level planning prefix before executable action generation.}
\label{fig:reasoning_representation}
\end{figure}

Given visual observations $o_t$, a language instruction $l$, and robot state $s_t$, our goal is to generate a short
  sequence of executable robot actions. Standard VLAs model this as direct action generation. Figure~\ref{fig:reasoning_representation}
  summarizes the main reasoning paradigms for VLA control, contrasting direct action generation with textual CoT, visual
  CoT, and our action-token CoT. In Coarse-to-Control, this action-token CoT is implemented as a planning stage followed
  by executable action generation:
\begin{align}
    z_t^{\mathrm{plan}} &\sim p_\theta(z^{\mathrm{plan}} \mid o_t,l,s_t), \\
    z_t^{\mathrm{exec}} &\sim p_\theta(z^{\mathrm{exec}} \mid o_t,l,s_t,z_t^{\mathrm{plan}}),
\end{align}
where $z_t^{\mathrm{plan}}$ denotes coarse action planning tokens and $z_t^{\mathrm{exec}}$ denotes executable action
tokens. During inference only $z_t^{\mathrm{exec}}$ is decoded and executed by the robot.
In our formulation, chain-of-thought takes the form of a coarse action
  trajectory that guides subsequent execution.
\subsection{Coarse-Fine Action Joint Tokenization}
\paragraph{Action Sub-resolution to Obtain Coarse Action}
To represent a coarse action plan, we summarize long-horizon future behavior into a shorter trajectory that captures stage-level motion intent rather than high-frequency executable control. Concretely, for each time step $t$ we take a long-horizon future action sequence $A_{t:t+H_p-1}$ and compress it into a coarse planning trajectory $\bar{A}_t$ with $K$ plan steps. Here $H_p$ is the number of source action timesteps, $K$ is the number of
  coarse plan chunks, and $c = H_p/K$ is the chunk size. For the relative action representation used by our
joint-mode tokenizer, this compression is implemented as
\begin{align}
    \bar{a}^{\mathrm{motion}}_{t+i}
  &= \sum_{j=0}^{c-1} a^{\mathrm{motion}}_{t+ic+j},\space \forall i \in \{i=0,\ldots,K-1\},\\
  \bar{a}^{\mathrm{gripper}}_{t+i}
  &= a^{\mathrm{gripper}}_{t+(i+1)c-1}, \space\forall i \in \{i=0,\ldots,K-1\}.
\end{align}
Each coarse plan step stores the net relative motion over one chunk and the final gripper state of that chunk. The
resulting $\bar{A}_t$ preserves the direction and stage-level intent of the future trajectory while discarding
high-frequency details; concrete settings are given in the implementation details. 

\paragraph{Dual-granularity Action Tokenizer}

We train a joint-mode residual-VQ action tokenizer with two modes: an execution mode $m=0$ and a planning mode $m=1$.
The execution mode tokenizes short-horizon action chunks $A_{t:t+H_e-1}$, while the planning mode tokenizes the coarse
long-horizon trajectory $\bar{A}_t$. Both modes share the same discrete action vocabulary, which encourages planning and execution tokens to remain in the same space.

Let $Q(\cdot,m)$ and $D(\cdot,m)$ be the mode-conditioned tokenizer and decoder. The target tokens are
\begin{align}
    z_t^{\mathrm{plan}} &= Q(\bar{A}_t, m=1), 
    \\
    z_t^{\mathrm{exec}} &= Q(A_{t:t+H_e-1}, m=0).
\end{align}
The tokenizer is trained jointly across the two modes with an action reconstruction objective
  and a residual-VQ regularization term that includes both codebook and commitment losses:
  \begin{equation}
      \mathcal{L}_{\mathrm{tok}} = \mathcal{L}_{\mathrm{rec}}^m + \mathcal{L}_{\mathrm{vq}}^m,
  \end{equation}
This objective encourages the planning branch to encode low-frequency long-horizon intent while keeping planning and
execution in the same control-aligned vocabulary. Here $H_p$ and $H_e$ denote the planning and executable horizons,
respectively. Additional tokenizer details and hyperparameters are provided in Appendix~\ref{app:joint-mode-action-tok}.

\subsection{Plan-Execute Autoregressive VLA}

After tokenization, the VLA is trained to predict the concatenated action suffix
$z_t = [z_t^{\mathrm{plan}}, z_t^{\mathrm{exec}}]$
conditioned on image tokens, language tokens, and proprioceptive state. The autoregressive factorization is
\begin{equation}
    p(z_t \mid o_t,l,s_t)=
  \prod_i p(z_{t,i}^{\mathrm{plan}} \mid x_t,z_{t,<i}^{\mathrm{plan}})
  \prod_j p(z_{t,j}^{\mathrm{exec}} \mid x_t,z_t^{\mathrm{plan}},z_{t,<j}^{\mathrm{exec}}),
\end{equation}
where $x_t=(o_t,l,s_t)$ is the multimodal context. The policy is trained with teacher-forced next-token prediction:
\begin{equation}
    \mathcal{L}_{\mathrm{VLA}} = -\sum_k \log p_\theta(z_{t,k}\mid x_t,z_{t,<k}).
\end{equation}

\begin{figure}[!htbp]
\centering
\includegraphics[width=0.9\linewidth]{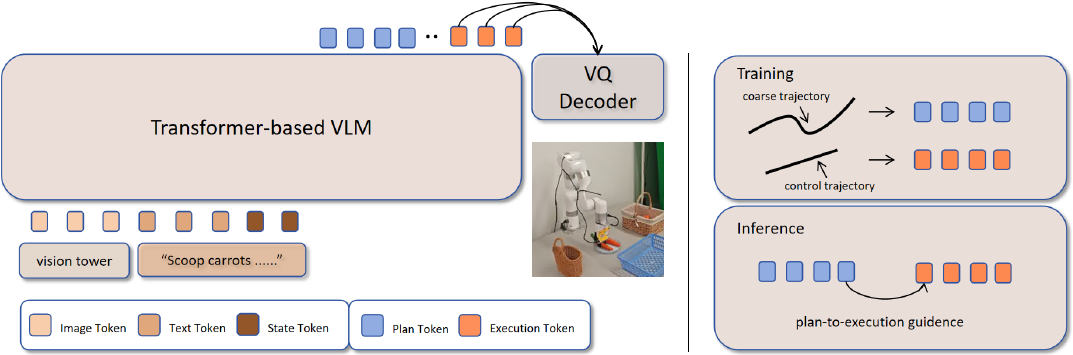}
\caption{Plan-execute VLA framework. The policy conditions on observations, language, and proprioception, predicts
coarse planning action tokens before executable action tokens, and decodes only the executable branch into continuous
robot actions.}
\label{fig:plan_execute_vla}
\end{figure}

At inference time, the model follows the same ordering: it first generates planning tokens and then executable tokens
conditioned on the generated plan. Only the executable tokens are decoded into robot actions, so planning remains an
internal context inside the autoregressive policy rather than a separate planner-controller interface.
Figure~\ref{fig:plan_execute_vla} shows the corresponding plan-then-execute flow.

\section{Experiments}

We evaluate Coarse-to-Control on simulation benchmarks and real-world robot manipulation tasks. Our experiments are
designed to address the following questions:
\begin{itemize}
  \item How effective is Coarse-to-Control compared with state-of-the-art methods on simulation benchmarks? (\S\ref{sec:main_simulation_experiments})
  \item How well does Coarse-to-Control perform on long-horizon real-world manipulation tasks? (\S\ref{sec:main_real_world_experiments})
  \item What additional advantages does action-token planning offer, and which design choices matter most? (\S\ref{sec:main_analysis})
\end{itemize}
Implementation details and extended ablations are provided in Appendix~\ref{app:implementation_details} and
Appendix~\ref{app:ablations}.

\subsection{Simulation Experiments}
\label{sec:main_simulation_experiments}

\paragraph{Benchmarks and Datasets.}
We evaluate on LIBERO~\citep{liu2023libero} and SimplerEnv-WidowX~\citep{li24simpler}. LIBERO contains four suites (Spatial, Object, Goal,
and Long), and we report per-suite and average success over 50 rollouts per task. SimplerEnv-WidowX evaluates real-to-sim
generalization on four WidowX manipulation tasks, and we report per-task and average success over 24 rollouts per task.
Dataset composition and training details are provided in Appendix~\ref{app:implementation_details}.

\paragraph{Baselines.}
We compare against recent VLA baselines spanning four reasoning styles: No-CoT, textual CoT, visual CoT, and Action CoT.
The full method lists for LIBERO and SimplerEnv-WidowX are reported in
Tables~\ref{tab:libero_benchmark_comparison} and~\ref{tab:simpler_widowx}. For SimplerEnv-WidowX, we follow the standard evaluation protocol. Implementation
details and training hyperparameters are provided in Appendix~\ref{app:implementation_details}.

\paragraph{Results.}
Tables~\ref{tab:libero_benchmark_comparison} and~\ref{tab:simpler_widowx} show that Coarse-to-Control achieves the
best overall performance on both benchmarks. These results
support the central claim of this paper: reasoning directly in the action-token space provides more control-aligned
guidance than explicit textual or visual CoT. 

\begin{table}[!htbp]
\centering
\scriptsize
\renewcommand{\arraystretch}{1.02}
\setlength{\tabcolsep}{3.0pt}
\caption{Performance comparisons with state-of-the-art methods on LIBERO, grouped by different CoT paradigms.}
\label{tab:libero_benchmark_comparison}
\begin{tabular}{@{}llccccc@{}}
\toprule
CoT Type & Method & Spatial $\uparrow$ & Object $\uparrow$ & Goal $\uparrow$ & Long $\uparrow$ & Overall $\uparrow$ \\
\midrule
No CoT & $\pi_0$-FAST~\citep{pertsch2025fast} & 96.4 & 96.8 & 88.6 & 60.2 & 85.5 \\
& SmolVLA~\citep{shukor2025smolvla} & 93.0 & 94.0 & 91.0 & 77.0 & 88.8 \\
& GR00T-N1~\citep{nvidia2025gr00tn1} & 94.4 & 97.6 & 93.0 & 90.6 & 93.9 \\
& $\pi_0$~\citep{black2024pi0} & 96.8 & 98.8 & 95.8 & 85.2 & 94.2 \\
& OpenVLA-OFT~\citep{kim2025openvlaoft} & 97.6 & 98.4 & 97.9 & 94.5 & 97.1 \\
\midrule
Textual CoT &  ThinkAct~\citep{huang2025thinkact} & 88.3 & 91.4 & 87.1 & 70.9 & 84.4 \\
& $\pi_{0.5}$~\citep{black2025pi05} & \textbf{98.8} & 98.2 & \textbf{98.0} & 92.4 & 96.8 \\
\midrule
Visual CoT & CoT-VLA-7B~\citep{zhao2025cotvla} & 87.5 & 91.6 & 87.6 & 69.0 & 81.1 \\
& WorldVLA~\citep{cen2025worldvla} & 87.6 & 96.2 & 83.4 & 60.0 & 81.8 \\
& DreamVLA~\citep{zhang2025dreamvla} & 97.5 & 94.0 & 89.5 & 89.5 & 92.6 \\
& UniVLA~\citep{wang2025univla} & 95.4 & 98.8 & 93.6 & 94.0 & 95.5 \\
& F1~\citep{lv2025f1} & 98.2 & 97.8 & 95.4 & 91.3 & 95.7 \\
& UD-VLA~\citep{chen2025udvla} & 94.1 & 95.7 & 91.2 & 89.6 & 92.7 \\
\midrule
Action CoT & MolmoAct-7B-D~\citep{duan2025molmoact} & 87.0 & 95.4 & 87.6 & 77.2 & 86.6 \\
& \textbf{Coarse-to-Control (ours)} & \textbf{98.8} & \textbf{100.0} & 97.8 & \textbf{95.0} & \textbf{97.9} \\
\bottomrule
\end{tabular}
\end{table}
\begin{table}[!htbp]
\centering
\scriptsize
\renewcommand{\arraystretch}{1.08}
\setlength{\tabcolsep}{3.2pt}
\caption{Performance comparisons with state-of-the-art methods on SimplerEnv-WidowX, grouped by different CoT paradigms.}
\label{tab:simpler_widowx}
\begin{tabular}{@{}llccccc@{}}
\toprule
CoT Type & Method & \shortstack{Put\\Spoon} & \shortstack{Put\\Carrot} & \shortstack{Stack\\Block} & \shortstack{Put\\Eggplant} & Overall $\uparrow$ \\
\midrule
No CoT & OpenVLA~\citep{kim2024openvla} & 0.0 & 0.0 & 0.0 & 4.1 & 1.0 \\
& Octo~\citep{ghosh2024octo} & 47.2 & 9.7 & 4.2 & 56.9 & 29.5 \\
& $\pi_0$~\citep{black2024pi0} & 29.1 & 0.0 & 16.7 & 62.5 & 40.1 \\
& CogACT~\citep{li2024cogact} & 71.7 & 50.8 & 15.0 & 67.5 & 51.3 \\
\midrule
Textual CoT & ThinkAct~\citep{huang2025thinkact} & 58.3 & 37.5 & 8.7 & 70.8 & 43.8 \\
\midrule
Visual CoT & F1~\citep{lv2025f1} & 50.0 & 70.8 & 50.0 & 66.7 & 59.4 \\
& UD-VLA~\citep{chen2025udvla} & 58.3 & 62.5 & 54.1 & 75.0 & 62.5 \\
\midrule
Action CoT & Coarse-to-Control (ours) & \textbf{100.0} & \textbf{95.8} & \textbf{79.2} & 58.3 & \textbf{83.3} \\
\bottomrule
\end{tabular}
\end{table}

\subsection{Real-World Experiments}
\label{sec:main_real_world_experiments}

\paragraph{Real-World Setup.}
We evaluate Coarse-to-Control on four physical manipulation tasks: putting the carrot on the plate, putting the
carrot on the plate and pressing the button, moving carrots from the plate into the basket, and clearing fruits and
vegetables from the table into the plate placed in the basket. All methods are trained on the same 50-demonstration-per-task
  budget, evaluated over 20 rollout trials under identical scene setups and
  success criteria. We compare the plan-based policy with faster, $\pi_0$, and $\pi_0$-fast variants.

\begin{figure}[t]
\centering
\includegraphics[width=\linewidth]{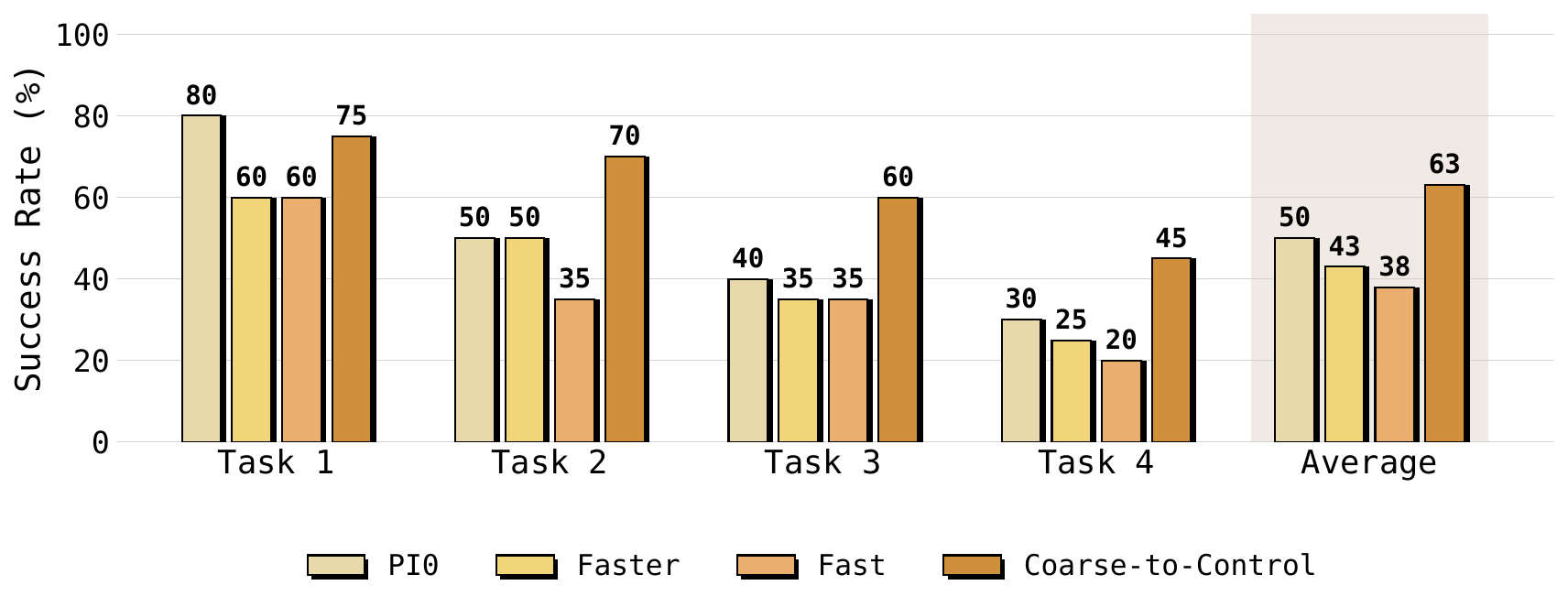}
\caption{Real-world task success rates (\%). We report final task success for each task and the average success
rate across four tasks. Carrot: put the carrot on the plate; Carrot+Button: put the carrot on the plate and press
the button; Plate$\rightarrow$Basket: move carrots from the plate into the basket; Cleanup: clear fruits and vegetables
from the table into the plate placed in the basket.}
\label{fig:real_world_results}
\end{figure}

\paragraph{Results.}
As shown in Figure~\ref{fig:real_world_results}, the plan-based policy achieves the highest average success rate among all
compared methods, reaching 62.5\% over four real-world tasks. It obtains the best performance on three out of four tasks. Although
$\pi_0$ performs best on the shortest single-stage carrot placement task, its advantage does not transfer to the longer
multi-stage settings. These results suggest that action-token planning improves robustness to error accumulation and
helps preserve temporal coordination over long-horizon physical manipulation. Subtask-level progress rates are reported
in Appendix~\ref{app:real_world_subtasks}.

\subsection{Analysis}
\label{sec:main_analysis}

We organize the analysis around four questions that explain the main results: how planning
horizon affects performance, why action-token plans are effective, whether planning and execution should share the same token space, and whether planning improves real-world long-horizon performance.

\begin{figure}[!htbp]
\centering
\begin{minipage}[t]{0.44\textwidth}
\centering
\footnotesize
\setlength{\tabcolsep}{3pt}
\renewcommand{\arraystretch}{1.02}
\refstepcounter{table}
\par\smallskip
\textbf{Table \thetable.} LIBERO planning-horizon $H_p$ ablation using joint-mode action-token planning.
\label{tab:horizon_ablation}
\par\smallskip
\resizebox{\linewidth}{!}{%
\begin{tabular}{lccccc}
\toprule
$H_p$ & Spatial $\uparrow$ & Object $\uparrow$ & Goal $\uparrow$ & Long $\uparrow$ & Overall $\uparrow$ \\
\midrule
0 & 97.00 & 99.60 & 95.00 & 94.20 & 96.45 \\
40 & 98.60 & 99.60 & 97.80 & 94.20 & 97.55 \\
160 & 98.80 & 100.00 & 97.80 & 95.00 & 97.90 \\
\bottomrule
\end{tabular}}
\end{minipage}
\hfill
\begin{minipage}[t]{0.54\textwidth}
\centering
\refstepcounter{figure}
\subcaptionbox{Attention: plan vs. w/o plan.
\label{fig:attention_comparison}}[0.58\linewidth]{%
    \includegraphics[height=1.5cm]{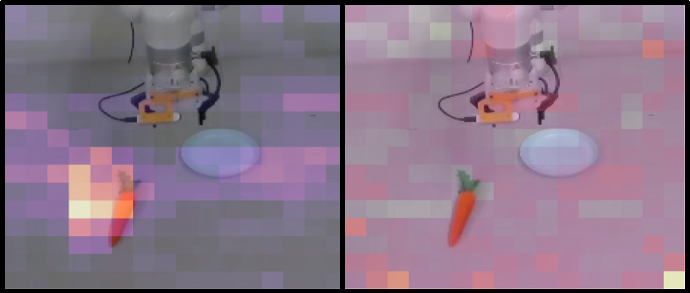}%
}
\hfill
\subcaptionbox{Plan visualization.
\label{fig:decoded_coarse_plan}}[0.4\linewidth]{%
    \includegraphics[height=1.5cm]{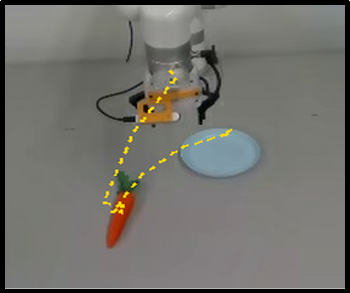}%
}
\par\smallskip
\footnotesize
\textbf{Figure \thefigure.} Qualitative analysis of joint-mode action-token planning.
\label{fig:qualitative_analysis}
\end{minipage}
\end{figure}

\paragraph{How Does Planning Horizon Matter?}
Table~\ref{tab:horizon_ablation} uses a fixed joint action tokenizer and varies only the planning horizon. First, planning itself is beneficial: increasing the planning horizon from 0 to 160 improves the LIBERO average success rate from 96.45\% to 97.90\%.
This interpretation is also consistent with Table~\ref{tab:joint_tokenizer_ablation}, where the Separate variant,
  which uses explicit planning tokens with two independent VQ tokenizers for planning and execution rather than a
  shared tokenizer vocabulary, still improves
  overall success from 95.40\% to 96.60\% relative to the no-plan Faster-AR baseline, which uses a single VQ tokenizer for execution only.
  This suggests that the gain does not come merely from the shared discrete action vocabulary. If shared tokenization were the sole driver, the $H_p=0$ variant would already recover most of the benefit. Instead, the gap shows that explicit plan
supervision helps the policy encode future task structure before action decoding, reducing the burden on the executable
branch to infer long-horizon intent from only the current observation. Second, once planning is introduced, the amount of
future context also matters. Increasing the planning source horizon from 40 to 160 improves the overall LIBERO score and
gives a consistent gain on the Long suite, showing that planning quality depends on access to sufficiently long future
context. The 40-step variant already shows that a short plan is better than no plan, but the 160-step variant performs
better still, indicating that a short plan can encode the next local maneuver whereas a longer source horizon better
captures stage transitions such as approach, grasp, transport, and placement. This is especially important when the
policy must choose actions that are locally valid yet only make sense in the context of a later subgoal.

\paragraph{Why Are Action-Token Plans Effective?}
Prior work suggests that chain-of-thought is effective not merely because it produces longer outputs, but because it introduces explicit intermediate states that transform difficult one-shot prediction into structured sequential computation~\citep{li2024cotserial,nye2021scratchpads,wang2023selfconsistency}. In embodied control, however, the benefit of CoT also depends on the representation medium. Textual CoT can provide high-level semantic decomposition but remains weakly constrained with respect to low-level motor behavior~\citep{zawalski2024ecot}, while visual CoT offers intuitive spatial structure yet requires generating long non-executable visual prefixes before action prediction~\citep{zhao2025cotvla}. By contrast, our method uses coarse action tokens as the intermediate state, so the reasoning trace lies closer to the control manifold and can align more directly with downstream executable actions. This coarse-to-fine design is also inspired by visual autoregressive modeling such as VAR, which first predicts global structure and then progressively refines local detail~\citep{tian2024var}.

To qualitatively inspect this mechanism, we use two diagnostics: attention
over image tokens and a decoded coarse trajectory from the predicted planning tokens. Figure~\ref{fig:attention_comparison}
compares the first-frame attention map produced with planning tokens against the no-plan baseline. With planning,
attention concentrates more strongly on task-relevant regions around the target object, gripper, and target area, whereas
the no-plan baseline is less anchored to the target-object interaction region. Figure~\ref{fig:decoded_coarse_plan}
shows that the decoded coarse plan already points toward the target before any executable actions are generated.
Together, these diagnostics suggest that the planning tokens carry object-target relations and future motion direction
  in a form that can guide subsequent control. In other words, the plan
branch appears to establish a grounded coarse intent before the execution branch predicts detailed motor commands, which
may help explain why action-space CoT can provide a more direct control signal for low-level control than
  higher-level textual or visual intermediates.

\paragraph{Should Planning and Execution Share Tokens?}
Table~\ref{tab:joint_tokenizer_ablation} isolates the tokenizer design. Faster-AR provides the no-plan reference, where
the policy directly predicts executable action tokens without an explicit planning prefix. Separate uses two independent
VQ action tokenizers for planning and execution, so the policy still predicts plan tokens before execution tokens but the
two branches no longer share a vocabulary. 
Joint-mode  improves overall success from 96.60\% to 97.90\% and raises the Long-suite score from 91.60\% to 95.00\%,
supporting the hypothesis that plan tokens should live on the same action-semantic manifold as executable tokens. When
the two branches use separate tokenizers, the execution policy must implicitly translate from a planning vocabulary into
an execution vocabulary, which weakens the conditioning signal. A shared token space reduces this interface mismatch and
makes the predicted plan easier to reuse as actionable guidance rather than as an abstract hint.
\begin{table}[!htbp]
\centering
\footnotesize
\setlength{\tabcolsep}{3.5pt}
\caption{Tokenizer-sharing ablation. Faster-AR is the no-plan baseline. Separate uses two independent VQ tokenizers for
planning and execution, while Joint-mode shares the vocabulary across the two branches.}
\label{tab:joint_tokenizer_ablation}
\begin{tabular}{lccccc}
\toprule
Tokenizer & Spatial $\uparrow$ & Object $\uparrow$ & Goal $\uparrow$ & Long $\uparrow$ & Overall $\uparrow$ \\
\midrule
Faster-AR & 99.40 & 98.80 & 94.80 & 88.60 & 95.40 \\
Separate & 97.40 & 99.60 & 97.80 & 91.60 & 96.60 \\
Joint-mode (ours) & 98.80 & 100.00 & 97.80 & 95.00 & 97.90 \\
\bottomrule
\end{tabular}
\end{table}
\paragraph{Does Planning Improve Real-World Long-Horizon Performance?}

\begin{wrapfigure}[11]{l}{0.46\textwidth}
\centering
\includegraphics[width=\linewidth]{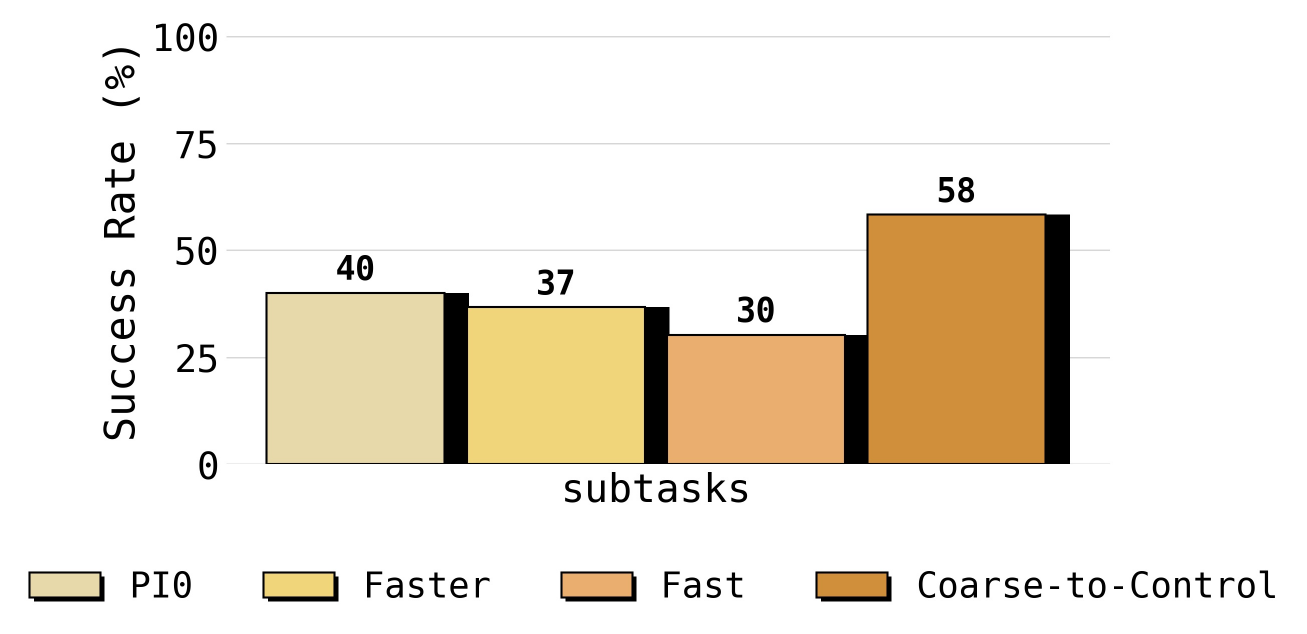}
\caption{Average success on the multi-stage real-world tasks.}
\label{fig:real_world_long_horizon}
\end{wrapfigure}

Figure~\ref{fig:real_world_long_horizon} summarizes average full-task success on the three multi-stage real-world tasks.
Coarse-to-Control achieves the highest long-horizon average and outperforms all baselines. This
suggests that the benefit of planning is especially pronounced when the robot must preserve task progress across multiple
stages instead of solving only a short single-step placement.
The subtask analysis in 
Table~\ref{tab:real_world_subtasks} further explains where this gain comes from. The plan-based
policy preserves more intermediate progress before final completion, suggesting that action-token plans help maintain the
full task trajectory under real-world perturbations, where a short-term action policy can complete early subtasks but fail
to finish the entire sequence. This effect is also consistent with the transfer probe in
Table~\ref{tab:libero_bridge_real_world}, where plan-based execution reduces the average successful rollout
  length from 436
  to 239 control steps. Fewer steps among successful trials suggest not only improved completion on long-horizon tasks but also
  improved execution efficiency among successful rollouts. This pattern is consistent with fewer local
  corrections,
  repeated re-approach behaviors, or inefficient detours before reaching the goal.

\FloatBarrier

\section{Limitations}

Although Coarse-to-Control shows strong performance in both simulation and real-world robot manipulation, it
  represents
  only one point in the broader design space of action-space reasoning. In particular, our current formulation
  realizes
  action-token chain-of-thought by predicting coarse future actions before execution, but how to construct more
  expressive
  and adaptive action-space reasoning schemes remains an open direction for future work. Moreover, while our
  results show
  that a joint action tokenizer can effectively align coarse planning actions and executable actions within a shared token
  space, how to more organically unify these two granularities and better capture shared structure in the
  action space
  remains to be explored.

\section{Conclusion}

We presented Coarse-to-Control, a plan-execute framework for Vision-Language-
  Action models that internalizes chain-of-thought reasoning into the action-token
  space. Rather than generating explicit textual rationales or visual subgoals,
  the policy first predicts a compact coarse action plan and then conditions
  executable action generation on this plan. A joint plan-execute tokenizer aligns
  long-horizon planning tokens and short-horizon executable tokens within a shared
  discrete action vocabulary, keeping planning close to the control manifold.
  Experiments on LIBERO, SimplerEnv-WidowX, and real-world robot manipulation
  tasks show that action-token reasoning improves benchmark performance,
  strengthens long-horizon robustness, and is especially effective on multi-stage
  tasks where intermediate progress must be preserved. More broadly, these results
  suggest that action-token planning provides a compact and promising alternative
  to explicit CoT generation for embodied policy learning without introducing a
  separate planner-controller interface.

\FloatBarrier
\bibliography{refs}

\clearpage
\appendix

\section*{Appendix}

\section{Additional Ablations}
\label{app:ablations}
\label{app:simulation_breakdown}

The main paper reports the core simulation, real-world, and design-ablation results. Here we provide additional
diagnostic breakdowns and implementation details that support the design choices.

\subsection{LIBERO Design Ablations}

\paragraph{Planning Horizon.}
Table~\ref{tab:horizon_ablation} reports completed LIBERO joint-mode runs for different planning source horizons.
Plan horizon 0 corresponds to the exec-only setting with no plan tokens, while 40 and 160 use increasingly longer coarse
plans. Increasing the planning source horizon from 40 to 160 improves average success and gives a gain on
\texttt{libero\_long}, indicating that a longer coarse plan helps on the LIBERO Long benchmark.

\paragraph{Joint Tokenizer versus Separate Tokenizers.}
Table~\ref{tab:joint_tokenizer_ablation} compares our joint-mode tokenizer with a non-joint variant that uses separate
tokenizers for coarse planning and executable actions. In the separate-tokenizer variant, long-horizon coarse plans and
short-horizon execution chunks are quantized by two independently trained codecs. The policy still predicts plan tokens
before execution tokens, but the two token streams no longer share a common codebook or token geometry. As a result, a
planning token is only an auxiliary latent hint: its token identity is not guaranteed to correspond to an executable
action pattern under the execution tokenizer. The autoregressive policy must therefore learn an additional
cross-tokenizer translation from the planning vocabulary to the execution vocabulary, which weakens the conditioning
signal and makes planning errors harder for the execution branch to interpret.

By contrast, the joint-mode tokenizer uses a shared action-token vocabulary with mode conditioning. This forces coarse
plans and executable chunks to live on the same action-semantic manifold, so a predicted planning token can more directly
bias the subsequent executable token distribution. The joint tokenizer improves average success from 96.60\% to 97.90\%,
with gains on LIBERO Long and the overall score.
A codebook-overlap diagnostic on re-encoded saved joint-mode reconstruction samples further shows partial but
nontrivial overlap between plan and execution token usage. Here the support-overlap mass is defined as
$\sum_i \min(p_i^{\mathrm{plan}}, p_i^{\mathrm{exec}})$ over the marginal plan-token and execution-token code
distributions, and is around 0.16 in our analysis, indicating shared token usage without collapse into the same
distribution.

\paragraph{Runtime Comparison Across Reasoning Media.}
For completeness, Table~\ref{tab:appendix_thinking_runtime} reports a controlled LIBERO runtime comparison between a CoT-VLA-style visual reasoning baseline and our action-token planning setup under the same rollout pipeline.

\par\smallskip
\refstepcounter{table}
\begin{center}
\footnotesize
\textbf{Table \thetable.} Controlled LIBERO runtime comparison between CoT-VLA-style visual reasoning and
Coarse-to-Control. Time is the average evaluation time per task.
\label{tab:appendix_thinking_runtime}
\par\smallskip
\begin{tabular}{lc}
\toprule
Method & Time (s) $\downarrow$ \\
\midrule
Visual CoT (CoT-VLA-style, 256 tok) & 3884.27 \\
Coarse-to-Control (42 tok) & 1325.53 \\
\bottomrule
\end{tabular}
\end{center}
\par\smallskip

The runtime gap is consistent with differences in autoregressive prefix length and reasoning medium. Our local
CoT-VLA-style baseline predicts a visual reasoning prefix before action tokens under the same rollout pipeline, so the
comparison keeps the evaluation procedure matched while varying the form of intermediate reasoning. In this setting, the
visual baseline uses a 256-token visual prefix, which substantially increases decoding cost before action generation. By
contrast, Coarse-to-Control uses 42 action-space guidance tokens in the LIBERO setting. This shorter control-aligned
prefix reduces decoding cost before action execution and avoids a separate image-to-action translation step.

\subsection{Real-World Subtask Progress}
\label{app:real_world_subtasks}

\begin{figure}[t]
\centering
\includegraphics[width=\linewidth]{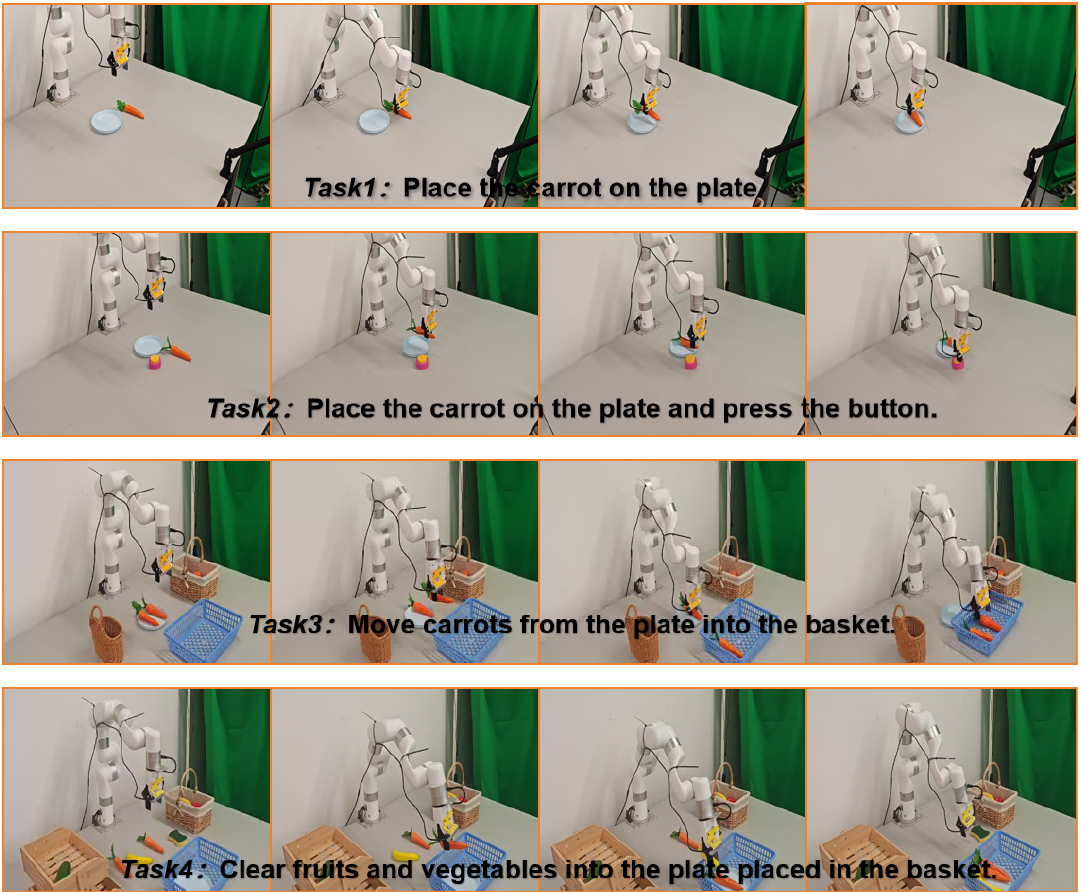}
\caption{Real-world setup and representative keyframes from the four physical manipulation tasks used in the real-world
evaluation.}
\label{fig:real_world_setup}
\end{figure}

Table~\ref{tab:real_world_subtasks} reports subtask-level progress rates for the multi-stage real-world tasks. Final
success is upper-bounded by earlier subtask completion, so the table diagnoses where failures enter the execution chain.
Plan-based execution preserves more progress from early subtasks to final completion, especially on the three multi-stage
tasks. Figure~\ref{fig:real_world_setup} provides representative keyframes for the four physical tasks used in
our real-world evaluation.

\begin{table}[!htbp]
\centering
\footnotesize
\setlength{\tabcolsep}{3.5pt}
\caption{Subtask progress rates (\%) in multi-stage real-world tasks. Final and subtask columns report the percentage
of rollouts that complete each corresponding stage.}
\label{tab:real_world_subtasks}
\begin{tabular}{llcccc}
\toprule
Method & Task & Final & Subtask 1 & Subtask 2 & Subtask 3 \\
\midrule
plan & Carrot+Button & 70.0 & 85.0 & 70.0 & -- \\
plan & Plate$\rightarrow$Basket & 60.0 & 70.0 & 60.0 & -- \\
plan & Cleanup & 45.0 & 70.0 & 55.0 & 45.0 \\
\midrule
faster & Carrot+Button & 50.0 & 75.0 & 50.0 & -- \\
faster & Plate$\rightarrow$Basket & 35.0 & 50.0 & 35.0 & -- \\
faster & Cleanup & 25.0 & 55.0 & 35.0 & 25.0 \\
\midrule
$\pi_0$ & Carrot+Button & 50.0 & 80.0 & 50.0 & -- \\
$\pi_0$ & Plate$\rightarrow$Basket & 40.0 & 60.0 & 40.0 & -- \\
$\pi_0$ & Cleanup & 30.0 & 55.0 & 40.0 & 30.0 \\
\midrule
$\pi_0$-fast & Carrot+Button & 35.0 & 40.0 & 35.0 & -- \\
$\pi_0$-fast & Plate$\rightarrow$Basket & 35.0 & 45.0 & 35.0 & -- \\
$\pi_0$-fast & Cleanup & 20.0 & 40.0 & 25.0 & 20.0 \\
\bottomrule
\end{tabular}
\end{table}

\subsection{Libero-Bridge Tokenizer Real-World Transfer}
\label{app:libero_bridge_real_world}

To disentangle the effect of action-token planning from tokenizer pretraining coverage, we additionally evaluate the
first two real-world tasks with tokenizers pretrained only on the LIBERO-Bridge mixture. This setting is not our main
real-world tokenizer setting: the main real-world experiments already compare plan and faster under the same downstream
real-world policy data, while using the broader real-world tokenizer described in
Table~\ref{tab:tokenizer_data_mixtures}. The LIBERO-Bridge tokenizer is narrower and more simulation/Bridge-oriented,
so it creates a stronger domain mismatch to our physical robot setup. We therefore use this setting as a controlled
transfer probe rather than as a recipe for maximizing absolute real-world performance. Under this probe, both plan and
faster variants use tokenizers pretrained on the same LIBERO-Bridge data mixture, while sharing the same downstream
policy data, robot platform, and evaluation protocol. The remaining comparison tests whether adding action-token plans
still helps when tokenizer pretraining coverage is matched but transfer to the physical robot remains imperfect.
Table~\ref{tab:libero_bridge_real_world} reports success rates as percentages over 20 trials per task, together with the
average number of rollout steps over successful trials.

\begin{table}[t]
\centering
\footnotesize
\setlength{\tabcolsep}{3.5pt}
\caption{Real-world transfer analysis using a tokenizer pretrained on the LIBERO-Bridge mixture. Success rates are
reported as percentages over 20 trials per task. Success Steps reports the average number of rollout steps over
successful trials only, rounded to the nearest control step.}
\label{tab:libero_bridge_real_world}
\begin{tabular}{lcccc}
\toprule
Method & Carrot Success $\uparrow$ & Carrot+Button Success $\uparrow$ & Avg. Success $\uparrow$ & Success Steps $\downarrow$ \\
\midrule
faster & 40.0 & 20.0 & 30.0 & 436 \\
plan & 70.0 & 35.0 & 52.5 & 239 \\
\bottomrule
\end{tabular}
\end{table}

Under the same LIBERO-Bridge tokenizer, plan-based execution improves success from 40.0\% to 70.0\% on Carrot
  and from
  20.0\% to 35.0\% on Carrot+Button, while reducing the average successful rollout length from 436 to 239
  control steps.
  This result is more consistent with a planning-mechanism gain than with a tokenizer-scaling gain. Even with a
  domain-mismatched tokenizer, the improvement suggests that plan tokens can still provide temporally extended
  guidance for
  real-world execution, rather than relying only on local action-only corrections. The reduction in successful
  rollout
  length is also consistent with more efficient successful trajectories, with fewer unnecessary corrective
  motions after
  success remains feasible.

\section{Implementation Details}
\label{app:implementation_details}

\subsection{Dataset Composition}
Table~\ref{tab:tokenizer_data_mixtures} summarizes the data used for action-tokenizer pretraining. For simulation
experiments, we use a compact mixture of LIBERO and Bridge. For real-world experiments, we use a broader
real-world-oriented mixture, following the same tokenizer-pretraining protocol but increasing the coverage of real robot
trajectories. Real-world task demonstrations are used for downstream policy training, while the pretrained action
tokenizer is kept fixed. For policy training, the input consists of primary and wrist camera views, language
instructions, and proprioceptive state. All datasets are stored in RLDS format.

\begin{table}[t]
\centering
\footnotesize
\setlength{\tabcolsep}{4pt}
\caption{Action-tokenizer pretraining data mixtures. The simulation tokenizer uses Libero and Bridge; the real-world
tokenizer uses a broader mixture of real-robot and robot-learning datasets.}
\label{tab:tokenizer_data_mixtures}
\begin{tabular}{llcl}
\toprule
Setting & Dataset & Weight & Usage \\
\midrule
Simulation & Libero~\citep{liu2023libero} & 5.0 & Simulation tokenizer \\
Simulation & Bridge~\citep{walke2023bridgedata2} & 1.0 & Simulation tokenizer \\
\midrule
Real world & Fractal~\citep{brohan2023rt2} & 1.0 & Real-world tokenizer \\
Real world & Kuka~\citep{kalashnikov2018qtopt} & 1.0 & Real-world tokenizer \\
Real world & Bridge~\citep{walke2023bridgedata2} & 1.0 & Real-world tokenizer \\
Real world & Droid(EEF)~\citep{khazatsky2024droid} & 1.0 & Real-world tokenizer \\
Real world & Libero~\citep{liu2023libero} & 5.0 & Real-world tokenizer \\
\bottomrule
\end{tabular}
\end{table}

\paragraph{Training setup.}
Table~\ref{tab:policy_training_setup} summarizes the downstream policy training setup. The LIBERO setting uses LIBERO
demonstrations, the SimplerEnv-WidowX setting uses Bridge demonstrations, and the real-world setting uses physical robot
demonstrations collected for each task. The action tokenizers are pretrained separately on 8$\times$H200 GPUs and kept
fixed during downstream VLA optimization.

\begin{table}[t]
\centering
\footnotesize
\setlength{\tabcolsep}{2.5pt}
\renewcommand{\arraystretch}{1.05}
\caption{Downstream policy training setup. All policies use AdamW with learning rate $2.5\times10^{-5}$, 1k warmup steps,
cosine decay, and weight decay $1\times10^{-10}$.}
\label{tab:policy_training_setup}
\begin{tabular}{lccc}
\toprule
Setting & Training data & Action-token horizon & Policy training \\
\midrule
LIBERO & LIBERO demonstrations & 20-step exec / 160-step plan & batch size 4, 60k steps \\
SimplerEnv-WidowX & Bridge demonstrations & 10-step exec / 80-step plan & batch size 16, 4 epochs \\
Real world & Physical robot demonstrations & 20-step exec / 160-step plan & batch size 8, 30k steps \\
\bottomrule
\end{tabular}
\end{table}

\subsection{Data Preprocessing}
For LIBERO policy training, we use a context window size equal to the policy condition length and a future action window
of horizon minus one. Images are resized to 224\,px, and the camera set contains the primary and wrist views. Action and
proprioceptive values are normalized with \texttt{q99}. The shuffle buffer size is 100k, and the datasets are sampled with
equal mixture weights.

For tokenizer pretraining, actions are normalized with \texttt{q99}. LIBERO and real-world policies use a 20-step
executable action horizon, while SimplerEnv-WidowX uses a 10-step executable horizon. The real-world tokenizer is trained
on the broader real-world-oriented mixture and then reused unchanged for downstream physical-robot policy training.

\subsection{Action Representation}
All LIBERO policies use 7-DoF end-effector actions. The executable action horizon is $H_e=20$ timesteps. For
Coarse-to-Control, the planning source horizon is $H_p=160$ timesteps in the main model, and the tokenizer compresses
this source window into $K=20$ coarse plan steps. This corresponds to a planning chunk size of 8. The 40-step ablation
uses the same 20-step executable horizon and $K=20$ plan steps, but compresses a 40-step source window with chunk size 2.
For the planning branch, motion dimensions are reduced with chunk-level summation, while the gripper dimension keeps the
last action in each chunk.

\subsection{Joint-Mode Action Tokenizer}
\label{app:joint-mode-action-tok}
The main model uses a joint-mode residual-VQ action tokenizer with two modes: execution mode and planning mode. Both
modes share the same discrete action vocabulary, and a learned mode condition specifies whether the tokenizer is encoding
short-horizon executable actions or coarse long-horizon planning actions. The tokenizer uses three residual VQ codebooks
with 4096 entries each. With a 20-step action horizon, two temporal patches, seven action dimensions, and three residual
codebooks, each branch is represented by $2 \times 7 \times 3 = 42$ action tokens in the LIBERO setting. The codec latent dimension is 64, the
action dimension is 7, and the mode-conditioning dimension is 64. The tokenizer reconstruction objective is an
action-space $\ell_1$ loss plus residual-VQ codebook and commitment losses, with commitment weight 0.25.

For completeness, the full tokenizer formulation used in the main model is as follows. Let
\begin{align}
  g_0(A_t) &= A_{t:t+H_e-1}, \\
  g_1(A_t) &= \bar{A}_t,
\end{align}
denote the mode-specific action preparation, where $g_1$ performs the planning-specific action sub-resolution by
chunking the $H_p$-step source trajectory and reducing each chunk into one coarse plan step, yielding the
$H_e$-step coarse plan representation. Let $P(\cdot)$ denote the shared codec
preprocessing applied after mode-specific preparation. In our implementation, $P(\cdot)$ corresponds to the patch-based
pretransform used by the residual-VQ autoencoder. Let $Q(\cdot,m)$ and $D(\cdot,m)$ be the mode-conditioned tokenizer
and decoder. The target tokens are
\begin{align}
  z_t^{\mathrm{plan}} &= Q(\bar{A}_t, m=1), \\
  z_t^{\mathrm{exec}} &= Q(A_{t:t+H_e-1}, m=0).
\end{align}

\begin{figure}[!htbp]
\centering
\includegraphics[width=\linewidth]{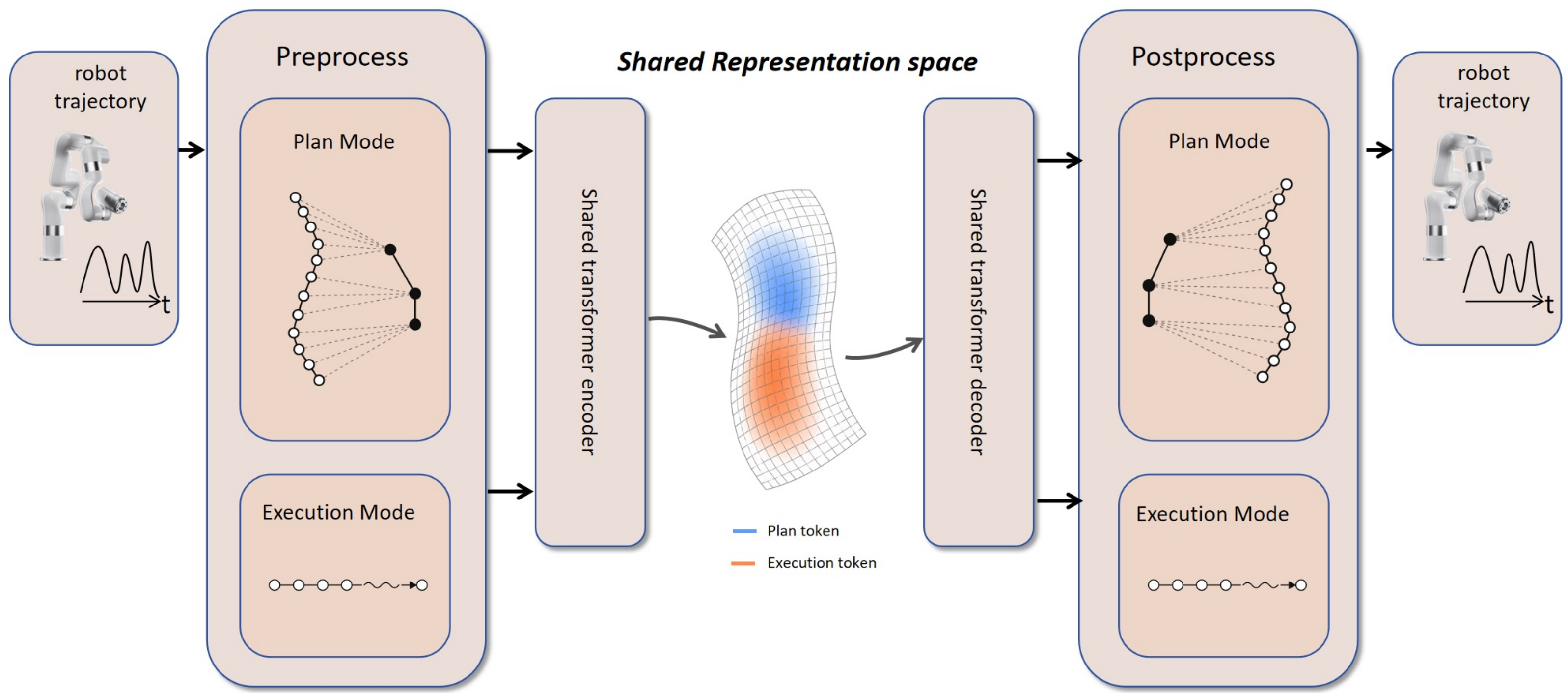}
\caption{Joint plan-execute action tokenizer. Execution mode represents short-horizon executable actions, planning
mode represents coarse long-horizon action-space guidance, and both modes share a discrete token vocabulary.}
\label{fig:joint_plan_execute_tokenizer}
\end{figure}

\paragraph{Joint Tokenizer Schematic.}
\label{app:joint_tokenizer_figure}
Figure~\ref{fig:joint_plan_execute_tokenizer} visualizes the two-mode tokenizer and the shared token space that aligns
planning and execution.

We define the preprocessed targets as
\begin{align}
  X_t^m &= P(g_m(A_t)), \\
  \hat{X}_t^m &= D(Q(X_t^m,m),m).
\end{align}
The reconstruction term is
\begin{equation}
  \mathcal{L}_{\mathrm{rec}}^m = \| X_t^m - \hat{X}_t^m \|_1.
\end{equation}
Let $e_t^{m,r}$ be the encoder residual entering the $r$-th codebook and $q_t^{m,r}$ be its selected code vector. With
$R_m$ the number of residual codebooks used in mode $m$, and $\mathrm{sg}(\cdot)$ denoting stop-gradient, the residual-VQ
terms are
\begin{align}
  \mathcal{L}_{\mathrm{commit}}^m
  &= \frac{1}{R_m}\sum_{r=1}^{R_m}\left\|e_t^{m,r}-\mathrm{sg}(q_t^{m,r})\right\|_2^2,\\
  \qquad
  \mathcal{L}_{\mathrm{codebook}}^m
  &= \frac{1}{R_m}\sum_{r=1}^{R_m}\left\|\mathrm{sg}(e_t^{m,r})-q_t^{m,r}\right\|_2^2.
\end{align}
For a training sample from mode $m$, the tokenizer objective is
\begin{equation}
  \mathcal{L}_{\mathrm{tok}}
  =
  \mathcal{L}_{\mathrm{rec}}^m
  + \mathcal{L}_{\mathrm{codebook}}^m
  + \beta\mathcal{L}_{\mathrm{commit}}^m
  ,
\end{equation}
where $\beta=0.25$ in our implementation. The tokenizer is trained on both execution and planning samples, with equal
mode sampling in our implementation.

\subsection{Separate-Tokenizer Ablation}
For the non-joint ablation in Table~\ref{tab:joint_tokenizer_ablation}, coarse planning and executable actions are
encoded by two independently trained VQ action tokenizers. The policy still generates planning tokens before execution
tokens, but the plan-token vocabulary and execution-token vocabulary are not shared. This setting tests whether the gain
comes merely from adding an action-level planning prefix, or from aligning planning and execution inside a common token
space.

\section{Training and Evaluation Details}
\label{app:training_eval_details}

\subsection{VLA Policy Training}
The policy uses a local PaliGemma-3B-based VLA backbone with a SigLIP visual encoder and the PaliGemma multimodal projector.
The multimodal prefix consists of image tokens, language tokens, and proprioceptive state. The autoregressive action
suffix is ordered as planning tokens followed by executable tokens. The policy is trained with teacher forcing over the
full plan-execute suffix. At inference time, planning tokens are generated first and used only as internal conditioning;
only executable tokens are decoded into continuous robot actions.

\subsection{Optimization}
Unless stated otherwise, policy training uses AdamW with cosine learning-rate decay, learning rate $2.5\times10^{-5}$,
warmup of 1{,}000 steps, weight decay $10^{-10}$, and maximum training length of 60k steps for LIBERO experiments. The
real-world experiments use the same optimizer family with the corresponding task-specific batch size and maximum
step budget defined in the training configuration.

\subsection{Model Configuration}
All locally trained policy variants use the same PaliGemma-based action-token VLA backbone. In our experiments, these models are initialized from $\pi_0$-FAST checkpoints. The VLA is a
unified multimodal model with separate VLM, proprioception, and action streams. In the evaluated configuration, the VLM
stream has hidden size 2048, the proprioception and action streams have hidden size 1024, the model uses 18 hidden
layers, 8 attention heads, 1 key-value head, head dimension 256, and maximum position length 8192. LoRA is disabled in
these evaluations. EMA weights are loaded for evaluation.

\subsection{LIBERO Evaluation}
We evaluate on \texttt{libero\_spatial}, \texttt{libero\_object}, \texttt{libero\_goal}, and \texttt{libero\_long}. The
evaluation uses 50 trials per task for the full benchmark summaries.  We report suite-level success rate and suite-level total evaluation time for all controlled comparisons in this
work.

\subsection{SimplerEnv-WidowX Evaluation}
For SimplerEnv-WidowX, we fine-tune on Bridge data with a single primary camera view, \texttt{q99} action normalization,
and image augmentation enabled. The policy uses a joint tokenizer with a 10-step executable horizon and an 80-step
planning horizon compressed into 10 coarse plan steps. We train for 4 epochs with batch size 16, AdamW, learning rate
$2.5\times10^{-5}$, 1k warmup steps, cosine decay, and weight decay $1\times10^{-10}$. Evaluation uses 24 trials per task,
loads EMA weights, and replans every 10 control steps.

\section{Real-World Evaluation Protocol}
\label{app:real_world_details}

The real-world benchmark contains four physical manipulation tasks: (1) putting the carrot on the plate, (2) putting the
carrot on the plate and pressing the button, (3) moving carrots from the plate into the basket, and (4) clearing fruits
and vegetables from the table into the plate placed in the basket. Each task is trained with 50 demonstrations and each method
is evaluated for 20 trials per task under the same scene setup and success criteria. The latter three tasks require
multi-stage progress; for these tasks we additionally report intermediate subtask completion rates in
Table~\ref{tab:real_world_subtasks}.

\paragraph{Success criteria.}
A rollout is treated as a final-task success only when the full language goal is completed. For multi-stage tasks, a
subtask is treated as completed when the corresponding intermediate object placement or interaction is completed before
timeout, even if a later stage fails. We report all real-world final and subtask results as percentages.

\end{document}